\documentclass[sn-mathphys,Numbered]{sn-jnl}% Math and Physical Sciences Reference Style
\usepackage{graphicx}%
\usepackage{multirow}%
\usepackage{amsmath,amssymb,amsfonts}%
\usepackage{amsthm}%
\usepackage{mathrsfs}%
\usepackage[title]{appendix}%
\usepackage{xcolor}%
\usepackage{textcomp}%
\usepackage{manyfoot}%
\usepackage{booktabs}%
\usepackage{algorithm}%
\usepackage{algorithmicx}%
\usepackage{algpseudocode}%
\usepackage{listings}%
\begin{document}

\title[]{Higher-order Motif-based Time Series Classification for Forced Oscillation Source Location in Power Grids}


\author[1]{\fnm{Long} \sur{Huo}}

\author*[1]{\fnm{Xin} \sur{Chen}}\email{xin.chen.nj@xjtu.edu.cn}

\affil[1]{\orgdiv{Center of Nanomaterials for Renewable Energy, State Key Laboratory of Electrical Insulation and Power Equipment, School of Electrical Engineering}, \orgname{Xi’an Jiaotong University}, \city{Xi'an}, \postcode{710054}, \state{Shaanxi}, \country{China}}


\abstract{Time series motifs are used for discovering higher-order structures of time series data. Based on time series motifs, the motif embedding correlation field (MECF) is proposed to characterize higher-order temporal structures of dynamical system time series. A MECF-based unsupervised learning approach is applied in locating the source of the forced oscillation (FO), a periodic disturbance that detrimentally impacts power grids. Locating the FO source is imperative for system stability. Compared with the Fourier analysis, the MECF-based unsupervised learning is applicable under various FO situations, including the single FO, FO with resonance, and multiple sources FOs. The MECF-based unsupervised learning is a data-driven approach without any prior knowledge requirement of system models or typologies. Tests on the UK high-voltage transmission grid illustrate the effectiveness of MECF-based unsupervised learning. In addition, the impacts of coupling strength and measurement noise on locating the FO source by the MECF-based unsupervised learning are investigated.}




\maketitle

\section{\label{sec:level1} Introduction}
For a wide range of spatiotemporal dynamical systems, time series of system states are essential for analyzing their inherent dynamics \cite{yuan2022single, severiano2021evolving, zeyringer2018designing}. Time series motifs are used for discovering higher-order structures of the time series of system dynamics and have proven effective for various tasks, from time series complexity measurement \cite{PhysRevE.87.022911} to machine malfunction detection \cite{9346704}. The time series motifs are smaller segments of the time series with fixed lengths and varying time delays, enabling to extract features of the time series at a mesoscopic scale. Motivated by the basic idea of time series motifs, a feature extraction approach called the motif embedding correlation field (MECF) is proposed by leveraging the correlations between pairs of time series motifs. The great potential of the proposed approach is demonstrated by the application in locating the forced oscillation (FO) source in power grids.

The FO is an essential phenomenon in nonlinear dynamical systems, and has attracts lots of research interests \cite{orazov2012forced, inoue2008nonlinear, han2023effects}. In power grids, the FO refers to an undesirable oscillation phenomenon resulting from exogenous periodic perturbations \cite{ghorbaniparvar2017survey}. Several FO events have already occurred in different regions, including the power grids of China \cite{ye2016analysis}, USA \cite{follum2016simultaneous}, and India \cite{mondal2019application}. The FO could sustain in the range of seconds to several hours, leading to detrimental impacts on the system's stability \cite{wang2017location}. To mitigate the consequence of FOs, it is critical to pinpoint the source of FOs, which is known as the FO source locating problem. Existing works address the FO source locating problem mainly in two ways: model-based methods and data-driven methods. The model-based methods such as energy functions \cite{tang2016research, maslennikov2020iso}, Luenberger observer \cite{li2018model}, transfer functions \cite{zhou2017locating}, and graph-theoretic approach \cite{nudell2015graph} take the information about the physical model of power grids into account. However, explicit model parameters or network typologies are not always available in practical applications \cite{wang2017location}. To overcome this limitation, it is more desirable to design data-driven FO source locating methods without resorting to the underlying physical model \cite{huang2018localization}.

In modern power grids, the growing adoption of monitoring technology, such as Phasor Measurement Units (PMUs), has led to increased interest in developing data-driven methods for locating the FO source \cite{usman2019applications}. Among the existing methods, machine learning has gained significant attention in recent years due to its powerful generalization ability \cite{meng2020time, chevalier2018bayesian, feng2022two, talukder2021low, matar2023transformer}. As a data-driven method, machine learning-based FO source location requires less or no information about the physical model of power grids. Nevertheless, the performance of machine learning-based methods often relies on effective feature selection and extraction \cite{ghorbaniparvar2017survey}. 

In addition, two common challenges need to be considered for locating the FO source.
Firstly, the FO can occur under the resonance condition, where the oscillation induced by the FO source resonates with a natural mode of the power system, rendering the FO source hard to be distinguished \cite{huang2020synchrophasor}. Secondly, when multiple concurrent FOs occur simultaneously at different locations in the power grid, all sources need to be accurately located.

The rest of the paper is organized as follows. In Sec. \ref{sec:level3}, the mathematical model and different situations of FOs in power grids are introduced. The MECF construction procedure of complex dynamical networks is given in Sec. \ref{sec:level2}. The MECF-based unsupervised learning approach for locating the FO source is proposed in Sec. \ref{sec:level2-2}. The numerical results of locating the FO by the MECF are presented in Sec. \ref{sec:level4}. Discussions are given in Sec.\ref{sec:level6} and conclusions are drawn in Sec. \ref{sec:level5}.

\section{ Transient dynamics of FO in power grids}\label{sec:level3}

The physical nature of a power grid can be coarse-grained and modeled as a coupled phase oscillator network \cite{anvari2020introduction, dorfler2013synchronization}. Each node in the network follows the dynamics derived from the swing equation in electrical power engineering \cite{choi2019synchronization}, $a,k,a$ the second-order Kuramoto model in physical communities \cite{filatrella2008analysis}:
\begin{equation}
\alpha _i\frac{d\omega _i}{dt}+\beta _i\omega _i=P_i-\sum_{j=1}^N{K_{ij}\sin \left( \delta _i-\delta _j \right)}
	                         \label{Kurmaoto}
\end{equation}
where $N$ represents the number of nodes in the network. The $i$th node ($i=1,2,\cdots, N$) is characterized by the power it generates ($P_i>0$) or consumes ($P_i<0$). Each node rotates at a phase angle $\delta _i\left( t \right)$ with an angular speed $\omega_i\left( t \right)$, both of which are measured under the reference frame $\varOmega =2\pi \times 50$Hz. The parameter $\alpha_i$ and $\beta_i$ is the inertia coefficient and damping constant of node $i$, respectively. The coupling strength between node $i$ and $j$ is represented by $K_{ij}$.

To maintain the stability of a power grid, the damping can dissipate small disturbances within the grid. However, sustained periodic disturbances such as FOs are sometimes difficult to be damped and can lead to instability problems \cite{kosterev1999model}. FOs originate from sinusoidal disturbances, which can be caused by cyclic loads, electrical oscillations due to controller malfunctions, or mechanical oscillations of generator turbines \cite{ye2016analysis}. The FO is mathematically described as:
\begin{equation}
\eta _i=\gamma_{i} e_i\cos \left( 2\pi \left( f_{i}t+\varphi_{i} \right) \right) + \xi _i
	         \label{disturbance}
\end{equation}

The $r.h.s.$ of Eq. \ref{disturbance} includes two terms. The first term represents a sinusoidal FO with $\gamma_{i}$, $f_{i}$, and $\varphi_{i}$, denoting the amplitude, frequency, and phase of the FO at node $i$. The binary parameter $e_i=1$ indicates that node $i$ is the source of the FO, while $e_i=0$ otherwise. The second term is a zero-mean Gaussian white noise $\xi _{i}\sim N(0,\sigma )$, where $\sigma$ is the standard deviation of the noise.

Linearizing Eq. \ref{Kurmaoto} and combining it with Eq. \ref{disturbance}, we can derive the linear stochastic power grid model under the FO as:
\begin{subequations}	
	\begin{equation}
\boldsymbol{\dot{Y}}=\boldsymbol{\varPhi Y}+\gamma \boldsymbol{e}\cos \left( ft+\varphi \right) +\boldsymbol{\xi }
	\label{linear1}
	\end{equation}
	\begin{equation}
\boldsymbol{\varPhi }=\left[ \begin{matrix}
	\vec{0}&		\vec{1}\\
	\boldsymbol{H}^{-1}\boldsymbol{L}&		-\boldsymbol{H}^{-1}\boldsymbol{D}\\
\end{matrix} \right] 
\label{linear2}
	\end{equation}
\label{linear3}
\end{subequations}
where $\boldsymbol{Y}=\left[\delta_1,\delta_2,\cdots,\delta_N,\omega_1,\omega_2,\cdots,\omega_N\right]^T$ is the vector representing system states of the power grid, $\boldsymbol{\xi}=\left[\xi_1,\xi_2,\cdots,\xi_N\right]^T$ is the vector of Gaussian white noise, and the matrices $\boldsymbol{H}$ and $\boldsymbol{D}$ are diagonal matrices of inertia coefficients and damping constants, respectively. The Laplacian matrix of the power grid is $\boldsymbol{L}$, $\vec{0}$, and $\vec{1}$ denotes the vector consisting of all zeros and all ones, respectively. The vector $\boldsymbol{e}$ indicates the source of the FO, $i.e.$, the $i$th element equals one if the $i$th node of the power grid is the source of FO, and all other elements are zero.

\section{\label{sec:level2} MECF for complex dynamical networks}
For a dynamical network system, the time series of system states at all nodes can be denoted as $\boldsymbol{X}\in \mathbb{R}^{N\times T}=\left\{ \boldsymbol{x}_i \right\}$, where $i=1,2,\cdots , N$ and $t=1,2,\cdots,T$ are the indices for nodes and discrete time points, respectively. Considering a power grid with N nodes, the time series of angular speed at node $i$ is denoted as $\boldsymbol{x}_i = [\omega_i(1), \omega_i(2), ..., \omega_i(T)]$. To construct the MECF of $\boldsymbol{x}_i$, the following procedure is adopted (where $\boldsymbol{x}_i$ is represented as $\boldsymbol{x}$ for simplicity).
First, the time delay embedding \cite{thiel2006spurious} is utilized to extract both dynamical and geometrical properties of $\boldsymbol{x}$ from a phase space perspective. The time delay embedding reconstructs the time series with embedding dimension $m$ and time delay $\tau$ as follows:

\begin{subequations}	
\begin{equation}
\boldsymbol{\hat{x}}=\left\{ \boldsymbol{\hat{x}}\left( t \right) ,t=1,2,\cdots T-\left( m-1 \right) \tau \right\} 
\end{equation}
\begin{equation}
\boldsymbol{\hat{x}}\left( t \right) =\sum_{j=1}^m{x\left( t+\left( j-1 \right) \tau \right)}\boldsymbol{e}_j
\end{equation}
\label{embedding}
\end{subequations}
where the vector $\boldsymbol{e}_j$ is an unit vector and span an orthogonal coordinate system. For example, if m=3 and $\tau$=2, then  
\begin{equation}
\boldsymbol{\hat{x}} = \begin{bmatrix}
  x(1)& x(3) & x(5) \\
  x(2)&  x(4)& x(6)\\
  x(3)&  x(5)& x(7)\\
  ...&  ...& ...\\
  x(T-4)&  x(T-2)& X(T)
\end{bmatrix}  
\end{equation}

Next, the sequence of time series motif of $\boldsymbol{\hat{x}}$ is defined as

\begin{subequations}	
	\begin{equation}
	\boldsymbol{M}_{d}^{n}=\left\{ \boldsymbol{M}_{d}^{n}\left( s \right) , s=1,2,\cdots ,T-\left( m-1 \right) \tau -\left( n-1 \right) d \right\} 
	\end{equation}	
	\begin{equation}
	\boldsymbol{M}_{d}^{n}\left( s \right) =\left\{ \boldsymbol{\hat{x}}\left( s \right) ,\boldsymbol{\hat{x}}\left( s+d \right) ,\boldsymbol{\hat{x}}\left( s+2d \right) ,\cdots ,\boldsymbol{\hat{x}}\left( s+\left( n-1 \right) d \right) \right\} 
	\end{equation}
	\label{motif}
\end{subequations}
where parameters $n$ and $d$ are the length of a time series motif and the displacement, respectively, $s$ is the start time point of a time series motif $M_{d}^{n}\left( s \right)$. For example, if n=3, d=2, m=3, and $\tau=2$, then for $s = T - (m-1)\tau - (n-1)d = T-8$, we have
\begin{equation}
\mathbf{M}_{d}^{n}(T-8)=\begin{bmatrix}
  x(T-8)& x(T-6) & x(T-4)\\
  x(T-6)&  x(T-4)& x(T-2)\\
  x(T-4)&  x(T-2)& x(T)
\end{bmatrix} 
\end{equation}

The motif $M_{d}^{n}\left( s \right)$ contains high-order local information of the original time series $\boldsymbol{x}$. Rather than analyzing the entire time series, investigating statistical relationships between different time series motifs allows us to find mesoscopic features. The sequence of statistical relationships between each pair of consecutive time series motifs $\boldsymbol{\tilde{M}}_{d}^{n}$ is defined as
\begin{equation}
\boldsymbol{\tilde{M}}_{d}^{n}=\left\{ \phi \left< \boldsymbol{M}_{d}^{n}\left( s \right) , \boldsymbol{M}_{d}^{n}\left( s+d \right) \right> , s=1,2,\cdots ,T-\left( m-1 \right) \tau - nd \right\}
\label{relationships} 	
\end{equation}
where the operation function $\phi \left< \cdot \right>$ calculates the statistical relationships between two time series motifs. The 2D correlation coefficient is employed as the operation function:
\begin{equation}
\phi \left< \boldsymbol{A},\boldsymbol{B} \right> =\frac{\sum_i{\sum_j{\left( \boldsymbol{A}_{ij}-\boldsymbol{\bar{A}} \right) \left( \boldsymbol{B}_{ij}-\boldsymbol{\bar{B}} \right)}}}{\sqrt{\left[ \sum_i{\sum_j{\left( \boldsymbol{A}_{ij}-\boldsymbol{\bar{A}} \right) ^2}} \right] \left[ \sum_i{\sum_j{\left( \boldsymbol{B}_{ij}-\boldsymbol{\bar{B}} \right) ^2}} \right]}}
\label{2Dcorrelation}
\end{equation}

In Eq. \ref{2Dcorrelation}, $\boldsymbol{A}$ and $\boldsymbol{B}$ denote two time series motifs, where $\boldsymbol{A}_{ij}$ and $\boldsymbol{B}_{ij}$ refer to the element at the $i$th row and $j$th column of $\boldsymbol{A}$ and $\boldsymbol{B}$, respectively. 
The mean of the elements in $\boldsymbol{A}$ and $\boldsymbol{B}$ are denoted as $\boldsymbol{\bar{A}}$ and $\boldsymbol{\bar{B}}$, respectively. 
The 2D correlation coefficient ranges from -1 to 1, where a value of 1 represents a strong positive correlation, -1 represents a strong negative correlation, and 0 indicates no correlation.
As shown in Eqs. \ref{relationships} and \ref{2Dcorrelation}, the correlation between two time series motifs not only varies with different starting time points $s$, but also with different displacements $d$, allowing to capture the multi-scale correlation between a pair of series motifs. Specifically, for small values of $d$, the short-range correlation between two close time series motifs is calculated, while for larger values of $d$, the long-range correlation between two distant time series motifs is calculated. 
The range of possible displacements is given by $1<d\leqslant d_{\max}$, yielding $1+\left \lfloor \frac{T-(m-1)\tau - (n+1)}{n-1} \right \rfloor$, where $T-(m-1)\tau > n$. 
Since the length of the sequence $\boldsymbol{\tilde{M}}_{d}^{n}$ defined in Eq. \ref{relationships} varies with different $d$, we construct a new sequence, denoted as $\boldsymbol{I}_{d}^{n}$, to align $\boldsymbol{\tilde{M}}_{d}^{n}$ to the same length. 
For each displacement $d$, $\boldsymbol{I}_{d}^{n}$ is initially set to be $\boldsymbol{\tilde{M}}_{d}^{n}$ and then appends with zero recursively until the total length of $\boldsymbol{I}_{d}^{n}$ equals to $2T-2\left( m-1 \right) \tau -n\left( 1+d_{\max} \right)$, which can be described as follows:
\begin{equation}
I_{d}^{n}\left( s \right) =\begin{cases}
	\phi \left< \boldsymbol{M}_{d}^{n}\left( s \right) , \boldsymbol{M}_{d}^{n}\left( s+d \right) \right> , if\,\,1<s\leqslant T-\left( m-1 \right) \tau -nd\\
	0,                              if\,\,T-\left( m-1 \right) \tau -nd<s\leqslant 2T-2\left( m-1 \right) \tau -n\left( 1+d_{\max} \right)\\
\end{cases}
\end{equation}

The stack of $\boldsymbol{I}_{d}^{n}$ with $d=1$ to $d=d_{\max}$ will form a matrix
\begin{equation}
\boldsymbol{G}^n=\left\{ \boldsymbol{I}_{1}^{n},\boldsymbol{I}_{2}^{n},\cdots ,\boldsymbol{I}_{d_{\max}}^{n} \right\}
\end{equation}

Finally, the following operation is conducted to generate the MECF
\begin{equation}
\boldsymbol{F}^n=\boldsymbol{G}^n+\boldsymbol{\grave{G}}^n
\label{MECF}
\end{equation}
where $\boldsymbol{\grave{G}}^n$ is the 180 degrees rotation of $\boldsymbol{G}^n$, and the matrix $\boldsymbol{F}^n$ is defined as the MECF of the time series $\boldsymbol{x}$. 

\begin{figure}[htb]
	\centering
	\includegraphics[width=0.95\textwidth]{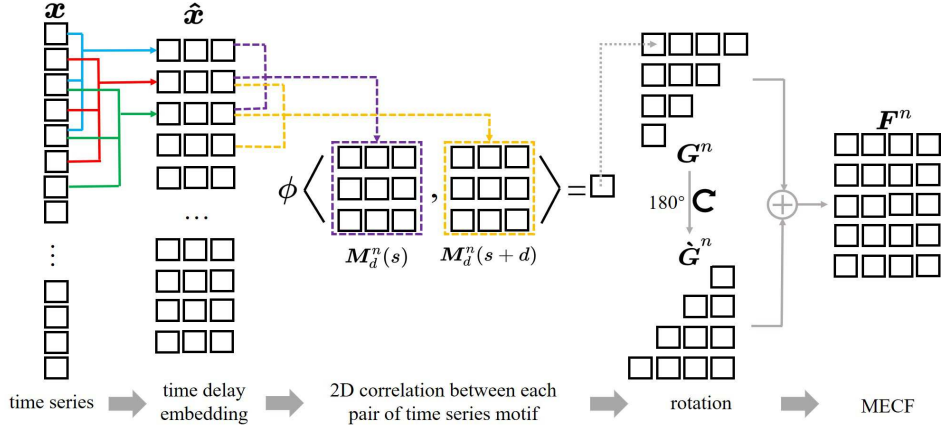}
	\caption{MECF construction procedure.}\label{fig11}
\end{figure}

In summary, given a time series $\boldsymbol{x}$, the corresponding MECF $\boldsymbol{F}^n$ can be constructed by three main procedures shown in Fig. \ref{fig11}. First, the time series $\boldsymbol{x}$ is reconstructed by the time delay embedding as $\boldsymbol{\hat{x}}$. Second, the sequence of time series motifs is defined, and the 2D correlation between each pair of time series motifs is calculated. Third, the matrix $\boldsymbol{G}^n$ recording the 2D correlations for all pairs of time series motifs is 180 degrees rotated as $\boldsymbol{\grave{G}}^n$. The MECF $\boldsymbol{F}^n$ is obtained as in Eq. \ref{MECF}.

\section{\label{sec:level2-2} MECF-based unsupervised learning for FO source location}

The FO source can be located by the MECF of each node in the power grid. The t-distributed stochastic neighbor embedding (t-SNE)\cite{van2008visualizing} is used to reduce the MECF of each node into a two-dimensional embedding. Then, the source of FO can be located by finding the outlier among the two-dimensional embeddings for MECFs of all nodes, $i.e.$, the embedding of the FO source is an outlier compared to the embeddings of other nodes.

An unsupervised clustering is employed to identify the outlier among the embeddings. First, a Euclidean distance matrix $\mathbf{L}$ is defined to measure the distance between each pair of embeddings. The matrix $\mathbf{L}$ is symmetrical, $i.e.$, $L_{i,j}=L_{j,i}$, and its element $L_{i,j}$ corresponds to the distance between the embeddings of node $i$ and node $j$, where $i,j=1,2,\cdots,N$ and $i\ne j$. Specifically, $L_{i,j}=\sqrt{\left( l_{i}^{x}-l_{j}^{x} \right) ^2+\left( l_{i}^{y}-l_{j}^{y} \right) ^2}$, where $l_{i}^{x}$ and $l_{j}^{x}$ are the coordinates of the embeddings of nodes $i$ and $j$, respectively.
Next, the average distance from the embedding of each node to the others is calculated as $\tilde{L}_i = \frac{1}{N-1} \sum_{j=1}^{N}L_{ij}$. All average distances $\tilde{L}_i$ belong to the average distance set $\mathcal{\tilde{L}}$, $i.e.$, $\tilde{L}_i\in\mathcal{\tilde{L}}$ for $i=1,2,...,N$.
The empirical distribution of all $\tilde{L}_i$ in the average distance set $\mathcal{\tilde{L}}$ is denoted as $p(\tilde{L}_i)$. By Chebyshev's inequality\cite{feller1967introduction}, which is applicable to arbitrary distributions provided they have a known finite mean and variance, it is known that a randomly selected sample $\tilde{L}_i$ from the average distance set $\mathcal{\tilde{L}}$ has a very low probability of approximately $4\%$ lying beyond five standard deviations of the mean of the distribution $p(\tilde{L}_i)$, denoted as $V_p$. The outlier is identified as any embedding with the average distance $\tilde{L}_i$ larger than $V_p$, and the corresponding FO source is located accordingly.

\section{\label{sec:level4} Numerical results}
\begin{figure}[htb]
	\centering
	\includegraphics[width=0.8\textwidth]{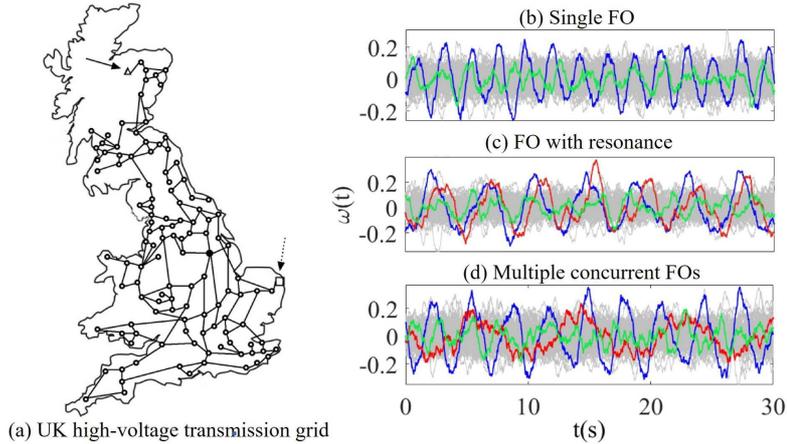}
	\caption{FOs in the UK high-voltage transmission grid. (a) The topology of the UK high-voltage transmission grid with 120 nodes and 165 edges. (b) Angular speeds of the $f=0.5Hz$ single FO. The curve in blue and green is the angular speed of the FO source and a randomly selected node, respectively. (c) Angular speeds of the $f=0.3Hz$ FO with resonance. The curve in blue, red, and green is the angular speed of the FO source, resonator, and randomly selected node, respectively. (d) Angular speeds of multiple concurrent FOs. The curve in blue, red, and green is the angular speed of the FO source with frequency $f=0.4Hz$, the FO source with frequency $f=0.2Hz$, and a randomly selected node, respectively. In (b) to (d), the curves in gray are the angular speeds of other nodes.}\label{fig1}
\end{figure}

The MECF-based unsupervised learning is applied for locating the FO source in the UK high-voltage transmission grid shown in Fig. \ref{fig1}. As shown in Fig.\ref{fig1}(a), the UK high-voltage transmission grid consists of 120 nodes and 165 edges \cite{manik2014supply, simonsen2008transient}.
The transient dynamics of each node is obtained by solving the stochastic power grid model in Eq. \ref{linear3}. 
The standard deviation of noise $\sigma=0.05$ and coupling strength $K_{ij}=K=15$. 
According to Eqs. \ref{embedding} to \ref{MECF}, the MECF is constructed for the time series of angular speed of each node. The embedding dimension m=3, the time delay $\tau=2$, the motif length of time series n=3, and the displacement d=2. For each node, the time series of angular speed from 0s to 30s are used for constructing the MECF, with a time step of 0.01. The following three different FO situations are considered to test the effectiveness of MECF-based unsupervised learning in locating the FO source:
\begin{itemize}
\item \textbf{Single FO}: Only one single source FO occurs in the power grid, which is the most common scenario in practical power systems.

\item \textbf{FO with resonance}: resonance is a well-known phenomenon in physics, where an applied periodic force with a frequency close to or equal to the natural frequency of the system will lead to an increased amplitude of oscillations. Resonance can occur when the FO frequency coincides with the natural frequency of a power grid's oscillation mode. Usually, the engineering intuition suggests that oscillations near the FO source would be most severe. However, it is counter-intuitive for the FO with resonance, $i.e.$, the oscillation amplitude at the resonator is even larger than that at the FO source. It is worth noting that the FO with resonance poses additional challenges in locating the source accurately, as the amplification of oscillation amplitudes at resonator nodes can lead to misleading information.

\item \textbf{Multiple concurrent FOs}: Multiple concurrent FOs involves the occurrence of more than one FO source simultaneously, which locate at different nodes in the power grid \cite{chevalier2018bayesian}. The amplitude and frequency of each FO source may be different, making it challenging to locate all the sources of multiple concurrent FOs.
\end{itemize}

Fig.\ref{fig1} (b) to (d) shows angular speeds under the single FO, FO with resonance, and multiple concurrent FOs. The single FO is shown in Fig.\ref{fig1}(b), where the curve in blue and green is the angular speed of the FO source with $f=0.5Hz$ and a randomly selected node, respectively. The angular speed of the FO source exhibits a more regular and periodic pattern in comparison to that of the randomly selected node. Fig.\ref{fig1}(c) illustrates the FO with resonance. The blue curve is the angular speed of the FO source with $f=0.3Hz$, while the red curve is the angular speed of the corresponding resonator. The curve in green is the angular speed of a randomly selected node. Although the geographical distance between the FO source and corresponding resonator is far, $i.e.$, as shown in Fig. \ref{fig1}(a), the FO source and corresponding resonator locates at the triangle node and black circle node, respectively, the amplitude and periodic pattern of the resonator are similar with the FO source. Fig.\ref{fig1}(d) shows the multiple concurrent FOs. The blue and red curves represent the angular speed of two FO sources with frequencies of $f=0.4$ Hz and $f=0.2$ Hz, respectively. The sources of these two FOs are situated at the triangle node and the square node as indicated in Fig.\ref{fig1}(a).

\begin{figure}
	\centering
	\includegraphics[width=1\textwidth]{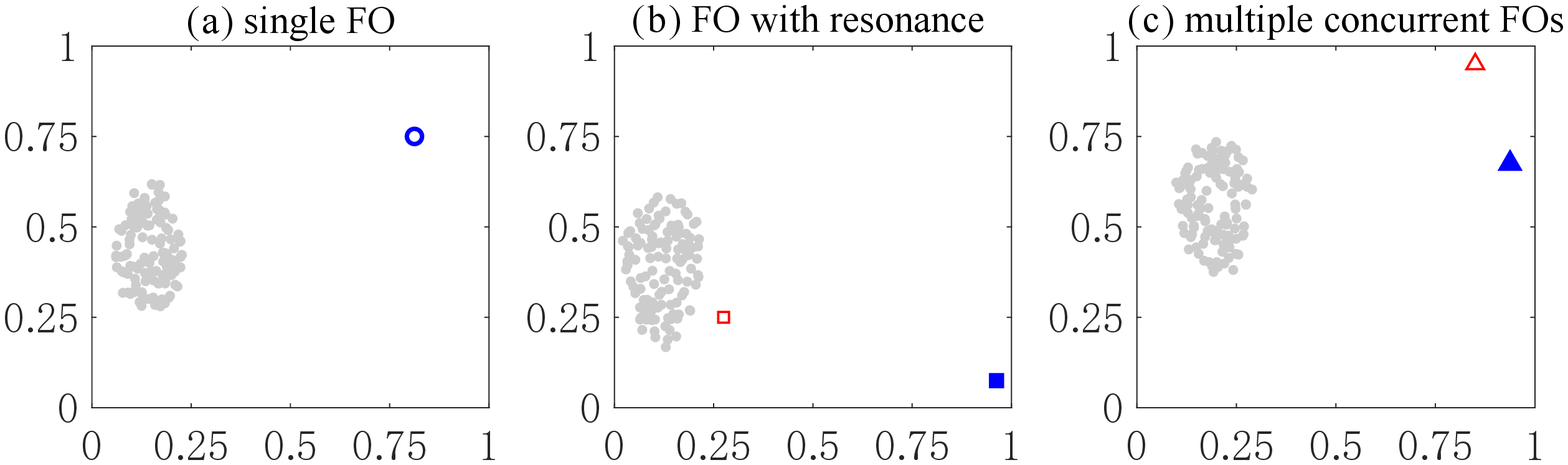}
	\caption{The t-SNE embeddings of MECFs. (a) Embeddings of the single FO. The blue circle indicates the embedding of the FO source. (b) Embeddings of the FO with resonance. The blue solid square and red square corresponds to the embedding of the FO source and resonator, respectively. (c) Embeddings of multiple concurrent FOs. The blue solid triangle and red triangle shows the embedding of the FO source with frequency 0.2 Hz and 0.4 Hz, respectively. The gray dots in (a) to (c) indicate the embeddings of other nodes excluding the FO source and resonator.}\label{fig3}
\end{figure}

Embeddings of the MECF are illustrated in Fig. \ref{fig3} (a) to (c), which are normalized within the range $[0, 1]$. 
Fig. \ref{fig3} (a) shows embeddings under the single FO, with the blue circle indicating the embedding of the FO source. 
Fig. \ref{fig3} (b) shows embeddings under the FO with resonance, where the blue solid square and the red square are the embeddings of the FO source and resonator, respectively. 
Embeddings of multiple concurrent FOs with frequencies of 0.2 Hz and 0.4 Hz are shown in Fig. \ref{fig3} (c), where the blue solid triangle and the red triangle represent the embedding of the FO source with the frequency of 0.2 Hz and 0.4 Hz, respectively. 
The gray solid points in Fig. \ref{fig3} (a) to (c) represent the embeddings of other nodes excluding the FO source and resonator.

From Fig. \ref{fig3} (a) to (c), it can be found that the embedding of the FO source is distinct from the highly-clustered embeddings of other nodes. 
In Fig. \ref{fig3} (a) to (c), the average distance $\tilde{L}_i$ of the FO source is 0.7, 0.85, and 0.75, respectively, while the average distance $\tilde{L}_i$ of other nodes is approximately 0.2, 0.25, and 0.22, respectively. 
In other words, the average distance $\tilde{L}_i$ of the FO source is at least three times greater than that of other nodes.
It is also noteworthy that, in Fig. \ref{fig3} (b), compared to the distance to the FO source, the embedding of the resonator is closer to the highly-clustered embeddings of other nodes. 

\begin{figure}[H]
	\centering
	\includegraphics[width=1.1\textwidth]{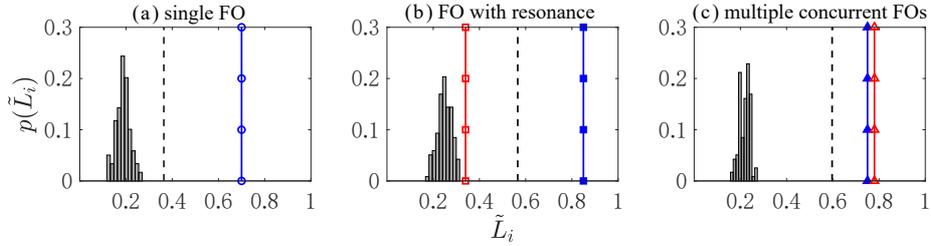}
	\caption{The empirical distributions $p(\tilde{L}_i)$ of average distances $\tilde{L}_i$. The gray bars in (a) to (c) represent the $p(\tilde{L}_i)$. The blue line with circles in (a), the blue line with squares in (b), and the blue and red lines with triangles in (c) indicate the $\tilde{L}_i$ of FO sources. The red line with squares in (b) represents the resonator. The black dashed lines indicate five standard deviations of the mean $V_p$ of distributions $p(\tilde{L}_i)$.}\label{fig3b}
\end{figure}

For t-SNE embeddings in Fig. \ref{fig3} (a) to (c),  the empirical distributions of average distances $p(\tilde{L}_i)$ are shown in Fig. \ref{fig3b} (a) to (c). The gray bars in Fig. \ref{fig3b} illustrate the probability distributions of $p(\tilde{L}_i)$. The $\tilde{L}_i$ corresponding to the FO sources are represented by the blue line with circles in Fig. \ref{fig3b} (a), the blue line with squares in Fig. \ref{fig3b} (b), and the blue and red lines with triangles in Fig. \ref{fig3b} (c).
The black dashed lines in Fig. \ref{fig3b} indicate the $V_p$. As introduced in Sec. \ref{sec:level2-2}, it is worth noting that the $\tilde{L}_i$ of the FO source is larger than the $V_p$, which can be treated as an outlier.
Furthermore, it can be found that the $\tilde{L}_i$ associated with the resonator, as illustrated by the red square line in Fig. \ref{fig3b} (b), is smaller than the $V_p$, $i.e.$, the $\tilde{L}_i$ of the resonator is not an outlier. Hence, the FO source can be located by finding the outlier even under the resonance condition.

\section{\label{sec:level6} Discussion}
\subsection{\label{sec:level6-1} Comparison: MECF-based unsupervised learning and Fourier analysis}
\begin{figure}
	\centering
	\includegraphics[width=1.05\textwidth]{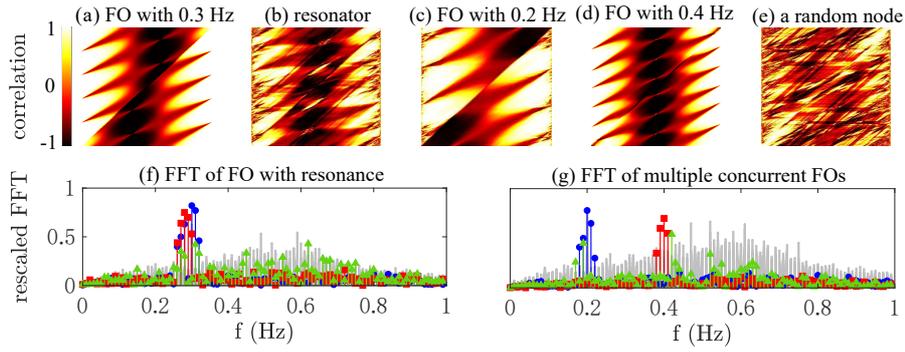}
	\caption{The MECF and Fourier analysis on locating the FO source in the UK high-voltage transmission grid. (a) The MECF of FO source with $f=0.3Hz$ and (b) the corresponding resonator. (c) and (d) The MECFs of multiple concurrent FOs sources with $f=0.2Hz$ and $f=0.4Hz$, respectively. (e) The MECF of a randomly selected node. (f) Fourier analysis of the FO with resonance. (g) Fourier analysis of multiple concurrent FOs. In (f), bars with blue dots, red squares, and green triangles correspond to the FO source with $f=0.3Hz$, the corresponding resonator, and a randomly selected node. In (g), bars with blue dots, red squares, and green triangles correspond to the FO source with $f=0.2Hz$, the FO source with $f=0.4Hz$, and a randomly selected node. The gray bars correspond to the other nodes excluding the FO source and resonator.}\label{fig2}
\end{figure}

The performance of the MECF and Fourier analysis on locating the FO source are compared. Fig.\ref{fig2} (a) to (d) shows visualizations of MECFs. The colors represent the correlations of time series motifs calculated by Eq.\ref{2Dcorrelation}, $i.e.$, the colors from black to white are correlations for pairs of time-series motifs from strongly negative to strongly positive. 
Fig.\ref{fig2} (a) is the visualization of MECF for the FO with the frequency $0.3Hz$, as shown by the blue curve in Fig. \ref{fig1} (c). The MECF shows alternatively bright and dark zones, reflecting that the MECF extracts periodic features from the angular speed time series of the FO source. As introduced in Sec. \ref{sec:level2}, for each element in the MECF, the correlation between a pair of time-series motifs is calculated. When a pair of time series motifs are both around the peaks or bottoms of the angular speed time series, the correlation exhibits a strong positive association, indicated by bright zones in Fig.\ref{fig2} (a). When one of the time series motifs is around the peak and the other is around the bottom of the angular speed time series, the correlation is characterized by a strong negative association, shown as dark zones in Fig.\ref{fig2} (a). 

For the FO with resonance, Fig.\ref{fig2} (b) is the visualization of the MECF of the resonator. In comparison to the MECF of the FO source in Fig.\ref{fig2} (a), the distinct boundaries between bright and dark zones become slightly stochastic, which is an obvious feature to distinguish the FO source from the corresponding resonator.
Fig. \ref{fig2} (c) and (d) show the MECF of FO source with frequency $0.2Hz$ and $0.4Hz$, respectively. In Fig. \ref{fig2} (c) and (d), the regular pattern of bright and dark zones can be found, while the number and size of the bright and dark zone are different.
Fig.\ref{fig2} (e) illustrates the MECF of a randomly selected node. The distribution of bright and dark zones is almost stochastic, making it distinct from MECFs of the FO and resonator.

The Fourier analysis, a standard signal processing method, is compared with the MECF on locating the FO source. Fig. \ref{fig2} (f) and Fig. \ref{fig2} (e) show the results of Fourier analysis. With the Fourier analysis, the FO source can be located by identifying peaks of Fourier components within the Fourier spectra.
Fig. \ref{fig2} (f) is the Fourier spectra of the FO situation shown in Fig. \ref{fig1} (c). The Fourier spectra have been re-scaled within the range [0, 1].
The bars with blue dots, red squares, and green triangles represent the Fourier components of the FO source, the resonator, and a randomly selected node, respectively.
The gray bars correspond to other nodes excluding the FO source and resonator. 
From the blue bars in Fig. \ref{fig2} (f), it can be found that the Fourier components for the FO source have a peak around $0.3Hz$, rendering it distinguishable from the Fourier components of other nodes. However, the peaks of the FO source and the corresponding resonator overlapped in the vicinity of $0.3Hz$. Hence, under the resonance condition, the FO source can't be located by the Fourier analysis.
For the Fourier spectra shown in Fig. \ref{fig2} (g), two peaks around $0.2Hz$ and $0.4Hz$ correspond to the frequencies of two different FO sources.  
The bars with blue dots, red squares, and green triangles are the Fourier components of the FO source with frequency $f=0.2Hz$, the FO source with frequency $f=0.4Hz$, and a randomly selected node, respectively. It can be found that the Fourier components of the randomly selected node are highly overlapped with the peaks of two FO sources. Hence, under multiple concurrent FOs, the FO source failed to be located by the Fourier analysis.

\subsection{\label{sec:level6-2} Impacts of coupling strength and noise}
The nature of FO in a power grid is a signal propagating on a spatiotemporal system. The propagation of signals are determined by the interplay between the network topology and the interaction dynamics \cite{hens2019spatiotemporal} . The topology of a power grid is weighted, $i.e.$, the coupling strength indicates the capacity of the transmission line between two nodes. Meanwhile, the measured data of practical power system dynamics such as the PMU data is inevitably accompanied by noise, and the noise will undermine the performance on locating the FO source\cite{khan2019separable}. 

\begin{figure}[H]
	\centering
	\includegraphics[width=1\textwidth]{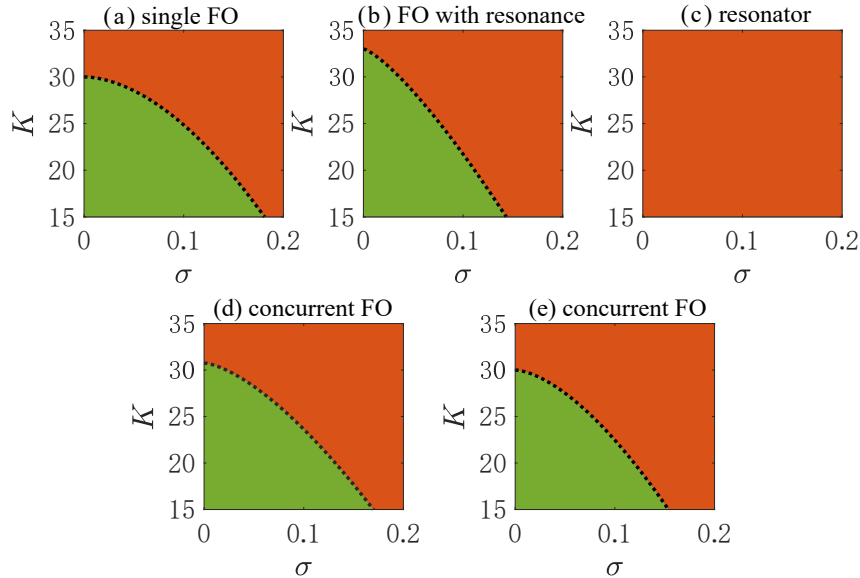}
	\caption{Locating the FO source under different coupling strength $K$ and noise standard deviation $\sigma$. The FO source can be correctly located in the power grid with $K$ and $\sigma$ within green areas. The FO source failed to be located in the power grid with $K$ and $\sigma$ within red areas. Black dashed lines at boundaries of green and red areas indicate the critical values of $K$ and $\sigma$. (a) and (b) corresponds to the single FO and the FO with resonance, respectively. (c) represents the resonator under the resonance condition. (d) and (e) illustrate multiple concurrent FOs. }\label{fig8}
\end{figure}

The impacts of coupling strength $K$ and standard deviation of noise $\sigma$ on locating the FO source by the MECF are shown in Fig. \ref{fig8}. To simplify the analysis, the homogeneous coupling strength is assumed throughout the network, denoted as $K_{ij}=K$ for all edges (i,j). As shown in Eq. \ref{disturbance}, the Gaussian white noise is added to the FO signal. The values of $K$ range from 15 to 30, while $\sigma$ ranges from 0 to 0.3. Fig. \ref{fig8} (a) to (c) correspond to the single FO, FO with resonance, and the corresponding resonator. 
Fig. \ref{fig8} (d) and (e) correspond to multiple concurrent FOs. 
Using t-SNE embeddings of MECFs to locate the FO source relies on the identification of outliers, $i.e.$, if the $\tilde{L}_i$ of an embedding exceeds the threshold $V_p$, it is an outlier and the corresponding node is the FO source.
In Fig. \ref{fig8}, values of $K$ and $\sigma$ in green regions ensure the $\tilde{L}_i$ of FO source is larger than $V_p$.
Conversely, values of $K$ and $\sigma$ in red regions result in the $\tilde{L}_i$ of FO source being smaller than $V_p$. Critical values of $K$ and $\sigma$ are represented by dashed black lines at boundaries between the green and red regions. For the power grid with $\sigma$ and $K$ smaller than critical values, the FO source can be correctly located by t-SNE embeddings of MECFs.
The critical value of coupling strength in Fig. \ref{fig8} (a), (b), (d), and (e) approximately ranges in $K \approx$ [15, 30]. According to existing literature \cite{rohden2014impact, rohden2012self, filatrella2008analysis}, the coupling strength in power systems typically ranges from $K$ = 10 to 22. Therefore, the MECF is adaptable for the typical range of coupling strength in power systems.
Meanwhile, both experimental tests and theoretical studies show that the noise standard deviation $\sigma$ in PMU data is usually $\sigma \approx$ [0, 0.1] \cite{brown2016characterizing, zhang2020experimental}. The critical value of noise standard deviation $\sigma$ in Fig. \ref{fig8} (a), (b), (d), and (e) ranges [0, 0.2], covering the range of noise standard deviation in experimental tests and theoretical studies. Hence, the MECF is adaptable for locating the FO source under measurement noise. 
In addition, it can be found that only the red region is shown for the resonator in Fig. \ref{fig8} (c). In other words, the $\tilde{L}_i$ of the resonator is consistently smaller than the $V_p$. Hence, for any coupling strength $K$ and noise standard deviation $\sigma$, the resonator has no impact on locating the FO source by the MECF.

\section{\label{sec:level5} Conclusions}
The MECF is proposed as a novel approach for extracting high-order structures in time series. The MECF reveals high-order structures of time series by analyzing correlations between time series motifs with different displacements and time delays. With the time series of transient dynamics in power grids, the MECF-based unsupervised learning is applied as a data-driven classification method for locating the FO source. Numerical results on the UK high-voltage transmission grid illustrate that the MECF-based unsupervised learning can correctly locate the FO source under situations of the single FO, the FO with resonance, and multiple concurrent FOs. Further discussions show that compared with the Fourier analysis, a standard signal processing approach, the MECF-based unsupervised learning has better performance in locating the FO source. Meanwhile, the MECF-based unsupervised learning is proved effective for a typical range of coupling strengths and measurement noise in power grids.

The utility of MECF is anticipated to extend beyond power grids to other dynamical systems with available data. In addition, the MECF can potentially act as a feature engineering tool to improve the classification and prediction performance of machine learning tasks, which will be our future studies.

\backmatter

\bmhead{Acknowledgments}
This work was supported in part by the HPC Platform, Xi’an Jiaotong University. Long Huo acknowledges the China Scholarship Council (CSC) scholarship.


\bibliography{sn-article.bib}% common bib file

\end{document}